\documentclass[10pt,twocolumn,letterpaper]{article}

\usepackage{cvpr}              
\usepackage[accsupp]{axessibility}

\usepackage{microtype}
\usepackage{xspace}
\usepackage{amsmath,amssymb,amsfonts}
\usepackage{amsthm}
\usepackage{mathtools}
\usepackage{booktabs}
\usepackage{multirow}
\usepackage{siunitx}
\usepackage{enumitem}
\setlist{leftmargin=*}
\usepackage{algorithm}
\usepackage{algpseudocode}
\usepackage{tikz}
\usetikzlibrary{arrows.meta,positioning,fit,calc,shapes.misc}
\usepackage{tabularx}
\usepackage[most]{tcolorbox}

\usepackage{longtable}
\usepackage{array}
\newcolumntype{L}[1]{>{\raggedright\arraybackslash}p{#1}}

\definecolor{clrNormal}{RGB}{39,125,61}
\definecolor{clrOsteopenia}{RGB}{180,120,20}
\definecolor{clrOsteoporosis}{RGB}{168,50,42}
\definecolor{clrBgNormal}{RGB}{235,248,239}
\definecolor{clrBgOsteopenia}{RGB}{255,248,230}
\definecolor{clrBgOsteoporosis}{RGB}{253,237,236}

\newtcolorbox{caseNormal}[1]{%
  enhanced, breakable=false, sharp corners=all,
  colback=clrBgNormal, colframe=clrNormal,
  boxrule=0.6pt, left=4pt, right=4pt, top=3pt, bottom=3pt,
  fonttitle=\bfseries\small, coltitle=white, colbacktitle=clrNormal,
  title={#1}}
\newtcolorbox{caseOsteopenia}[1]{%
  enhanced, breakable=false, sharp corners=all,
  colback=clrBgOsteopenia, colframe=clrOsteopenia,
  boxrule=0.6pt, left=4pt, right=4pt, top=3pt, bottom=3pt,
  fonttitle=\bfseries\small, coltitle=white, colbacktitle=clrOsteopenia,
  title={#1}}
\newtcolorbox{caseOsteoporosis}[1]{%
  enhanced, breakable=false, sharp corners=all,
  colback=clrBgOsteoporosis, colframe=clrOsteoporosis,
  boxrule=0.6pt, left=4pt, right=4pt, top=3pt, bottom=3pt,
  fonttitle=\bfseries\small, coltitle=white, colbacktitle=clrOsteoporosis,
  title={#1}}

\definecolor{cvprblue}{rgb}{0.21,0.49,0.74}
\usepackage[pagebackref,breaklinks,colorlinks,allcolors=cvprblue]{hyperref}

\makeatletter
\providecommand{\theHalgorithm}{\arabic{algorithm}}
\providecommand{\theHALG@line}{\theHalgorithm.\arabic{ALG@line}}
\makeatother

\def\paperID{*****} 
\def\confName{CVPRW}
\def\confYear{2026}

\newcommand{\protomedx}{ProtoMedX\xspace}
\newcommand{\protomedagent}{ProtoMedAgent\xspace}

\theoremstyle{definition}

\theoremstyle{plain}

\title{ProtoMedAgent: Multimodal Clinical Interpretability via Privacy-Aware Agentic Workflows}

\author{
Alvaro Lopez Pellicer$^{1,}$\footnotemark[2], Plamen Angelov$^{1}$, Marwan Bukhari$^{2}$, Yi Li$^{1}$, Eduardo Soares$^{3}$, Jemma Kerns$^{2}$ \\[1.5mm]
$^{1}$School of Computing and Communications, Lancaster University \\
$^{2}$Lancaster Medical School, Lancaster University \qquad $^{3}$PUC-Rio, Puc-Behring Institute for AI
}

\begin{document}
\maketitle

\renewcommand{\thefootnote}{\fnsymbol{footnote}}
\footnotetext[2]{Corresponding author. Email: a.lopezpellicer@lancaster.ac.uk}
\renewcommand{\thefootnote}{\arabic{footnote}}

\begin{abstract}
While interpretable prototype networks offer compelling case-based reasoning for clinical diagnostics, their raw continuous outputs lack the semantic structure required for medical documentation. Bridging this gap via standard Retrieval-Augmented Generation (RAG) routinely triggers ``retrieval sycophancy,'' where Large Language Models (LLMs) hallucinate post-hoc rationalizations to align with visual predictions. We introduce ProtoMedAgent, a framework that formalizes multimodal clinical reporting as an iterative, zero-gradient test-time optimization problem over a strict neuro-symbolic bottleneck. Operating on a frozen prototype backbone, we distill latent visual and tabular features into a discrete semantic memory. Online generation is strictly constrained by exact set-theoretic differentials and a reflective Scribe-Critic loop, mathematically precluding unsupported narrative claims. To safely bound data disclosure, we introduce a semantic privacy gate governed by $k$-anonymity and $\ell$-diversity. Evaluated on a 4,160-patient clinical cohort, ProtoMedAgent achieves 91.2\% Comparison Set Faithfulness where it fundamentally outperforms standard RAG (46.2\%). ProtoMedAgent additionally leverages a binding $\ell$-diversity phase transition to systematically reduce artifact-level membership inference risks by an absolute 9.8\%.
\end{abstract}

\section{Introduction}

In clinical practice, high-stakes diagnostic assessment is inherently multimodal. For instance, in bone health screening, DEXA imaging provides dense structural and spatial cues, while tabular patient variables (e.g., age, BMI, FRAX risk factors) contextualize the overarching disease severity \cite{trace, Pellicer_2025_ICCV}. While modern deep networks readily optimize over these heterogeneous modalities to achieve high predictive accuracy, they leave a critical gap in clinician-facing documentation: they fundamentally fail to generate transparent, auditable, and privacy-compliant rationales. This creates a mismatch between model performance and clinical usability, where accurate predictions lack trustworthy justification. This need is increasingly regulatory as well as technical: emerging EU rules reward clinical AI designs that are auditable, human-overseeable, robust, and compatible with controlled data access and interoperable data infrastructures~\cite{eu_ai_act_2024,eu_data_act_2023}.

\begin{figure}[t]
    \centering
    \includegraphics[
        width=\columnwidth,
        height=0.42\textheight,
        keepaspectratio
    ]{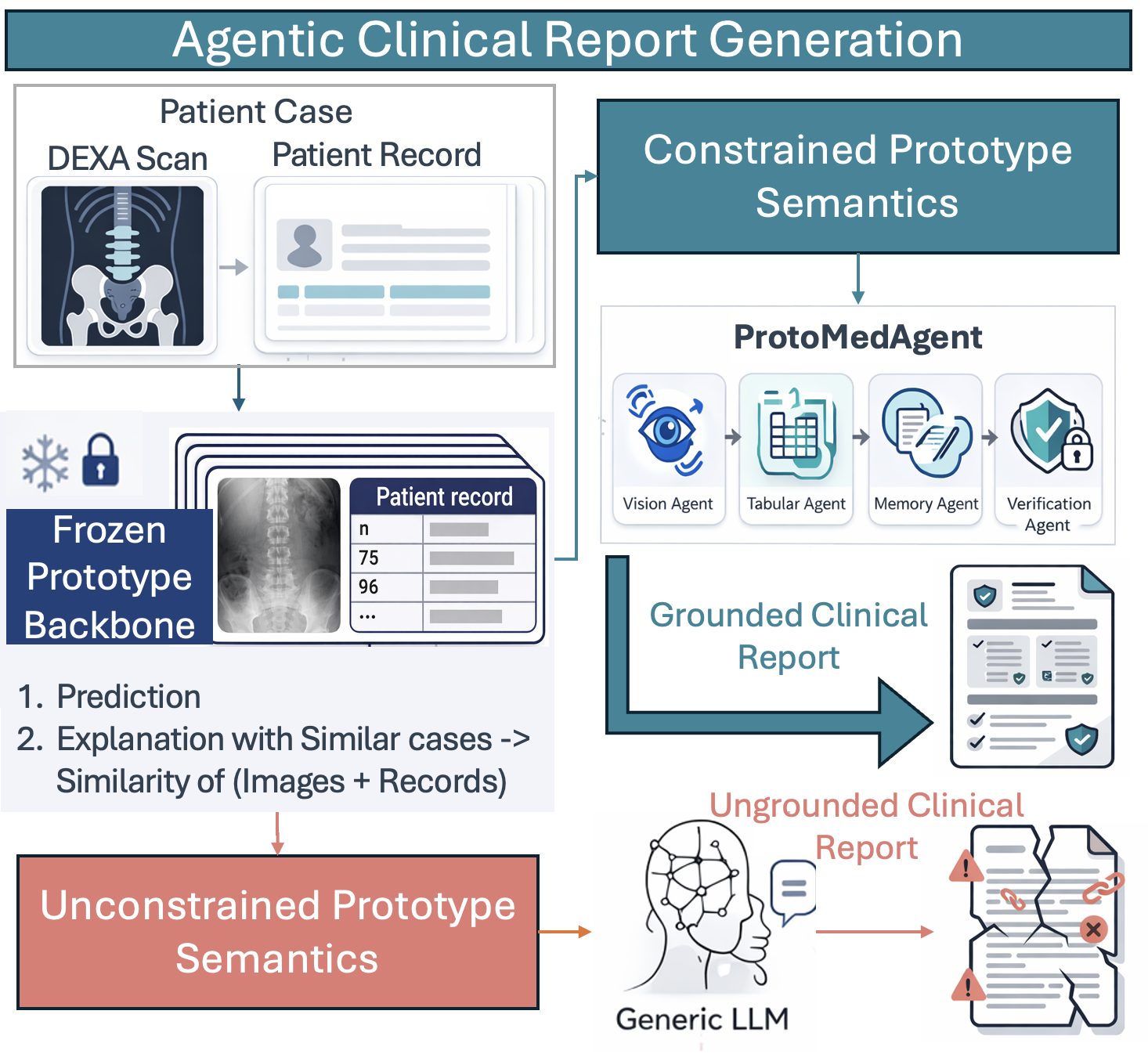}
    \caption{\textbf{Overview of an Agentic Clinical Report Generation framework} A multimodal patient case, comprising a lumbar DEXA scan and a patient record, is first processed by a frozen prototype backbone that retrieves raw visual exemplars and tabular statistics. Reconciling these outputs through unconstrained prototype semantics via a generic LLM yields an ungrounded, hallucination-prone clinical report. In contrast, by enforcing constrained prototype semantics, \protomedagent\ utilizes a suite of dedicated agents (Vision, Tabular, Memory, and Verification) to safely translate retrieved evidence into a fully grounded and verifiable clinical report.}
    \label{fig:protomedagent_intro}
\end{figure}

Interpretable-by-design prototype networks have emerged as a compelling alternative, offering case-based reasoning by exposing training exemplars with the logic of ``this looks like that''~\cite{chen2019protopnet, nauta2023pipnet, nauta2021prototree}. Yet, the raw evidence surface of prototype retrieval---comprising scalar similarity scores and continuous feature maps---is semantically disconnected from clinical reporting workflows. The standard intervention of coupling a Large Language Model (LLM) via Retrieval-Augmented Generation (RAG) to interpret these latent prototypes routinely collapses. When tasked with reconciling unconstrained visual features and clinical priors, LLMs exhibit severe multimodal hallucination and retrieval sycophancy, fabricating similarities to post-rationalize the network's prediction~\cite{sharma2023sycophancy, rrv2024chaos}.

Inspired by recent advances in reflective agentic systems, we introduce \protomedagent\ (Figure \ref{fig:protomedagent_intro}), a framework that bridges the gap between continuous visual reasoning and discrete clinical documentation. We hypothesize that faithful multimodal reporting requires decoupling the \emph{diagnostic forward pass} from the \emph{evidence documentation}. Rather than treating reporting as an unconstrained generative pass, we cast it as an \textbf{iterative LLM-in-the-loop test-time optimization problem} operating over a strict neuro-symbolic bottleneck.

Our framework acts upon a frozen prototype backbone, \protomedx~\cite{Pellicer_2025_ICCV}. Offline, we project the continuous prototype space into a privacy-gated semantic memory, distilling visual and tabular distributions into discrete assertions bounded by strict $k$-anonymity and $\ell$-diversity. Online, we introduce a reflective refinement loop: exact set-theoretic differentials are computed between the query and the retrieved memory, establishing an absolute admissible state space. A Scribe-Critic loop then iteratively optimizes the generated report, utilizing the LLM as both a proposal distribution and a constraint evaluator until the artifact achieves perfect neuro-symbolic alignment.

This work investigates the following research questions:
\begin{itemize}
    \item \textbf{RQ1 (Neuro-Symbolic Grounding):} How can latent multimodal prototype retrieval be converted into a discrete semantic bottleneck that promotes evidence-grounded generation and reduces the space for unsupported LLM outputs?
    \item \textbf{RQ2 (Test-Time Optimization):} Can an iterative reflective critic at test time improve reporting faithfulness without requiring gradient updates or altering the predictive behavior of the frozen vision backbone?
\end{itemize}

\paragraph{Contributions.}
\begin{enumerate}
    \item \textbf{Agentic LLM-in-the-Loop Test-Time Optimization:} We formalize clinical reporting as a zero-gradient test-time optimization process, leveraging an offline privacy-gated memory and an online reflective reasoning loop to safely decode a frozen prototype classifier.
    \item \textbf{Neuro-Symbolic Verification:} We replace heuristic LLM comparisons with deterministic set-theoretic differentials. This is enforced via an iterative Scribe-Critic optimization loop that guarantees citation accuracy and typed-value consistency, fundamentally mitigating retrieval sycophancy.
    \item \textbf{Provable Privacy-Utility Bounds:} We introduce a novel semantic release gate utilizing $k$-anonymity and $\ell$-diversity for continuous prototype memory, demonstrating mathematically bounded disclosure risk and improved empirical faithfulness on a 4,160-patient clinical cohort.
\end{enumerate}


\section{Related Work}
\label{sec:related_work}

Interpretable-by-design prototype models natively explain their predictions by retrieving latent exemplars from the training distribution, operationalizing the intuitive logic of ``this looks like that''~\cite{chen2019protopnet, nauta2023pipnet, nauta2021prototree, angelov2020xdnn}. In high-stakes medical domains, such inherently transparent architectures are strongly preferred over post-hoc explainers~\cite{rudin2019stop}. While recent approaches like \protomedx~\cite{Pellicer_2025_ICCV} advance this paradigm via input-dependent multimodal gating, the native evidence surface of these models---comprising scalar similarities and continuous feature activations---exhibits a profound semantic gap when mapped to discrete clinical reporting workflows. Recent works have explored the use of LLMs for clinical report generation and multimodal reasoning over medical images and structured data~\cite{nam2025multimodal}, including radiology report generation~\cite{deperrois2025radvlm} and clinical decision support systems \cite{singhal2023large}. However, these approaches typically rely on end-to-end generative pipelines without explicit or verifiable grounding to underlying evidence, limiting their reliability in high-stakes settings~\cite{artsi2025large, singhal2023large, ji2023survey}.

To bridge this representational gap, recent advancements have increasingly scaled test-time compute by configuring Large Language Models (LLMs) as compound, reflective agents. Frameworks such as AlphaCodium~\cite{ridnik2024alphacodium} and TimeXL~\cite{jiang2025timexl} demonstrate that interleaving generation with iterative self-critique~\cite{yao2022react, shinn2023reflexion} yields significant performance gains for complex, multi-step reasoning without requiring gradient updates. However, simply grounding these agentic loops via standard Retrieval-Augmented Generation (RAG)~\cite{lewis2020rag} introduces critical failure modes in clinical settings. Tasking an LLM to heuristically \emph{compare} retrieved items reliably induces retrieval sycophancy and multimodal hallucination~\cite{sharma2023sycophancy, rrv2024chaos, malmqvist2024sycophancy}. Because language models tend to prioritize internal semantic priors over contradicting external evidence, even cited RAG outputs frequently suffer from spurious post-rationalization~\cite{wallat2025correctness, niu2024ragtruth}. 

Modern constrained decoding engines, including Outlines~\cite{willard2023outlines}, XGrammar~\cite{dong2024xgrammar}, and JSONFormer~\cite{jsonformer2023}, attempt to reign in this behavior by enforcing strict syntactic schema adherence~\cite{geng2025jsonschemabench, shorten2024structuredrag}. Yet, while these tools guarantee structural validity, they cannot mathematically guarantee semantic faithfulness. Resolving this tension requires replacing unconstrained heuristic comparison with deterministic, neuro-symbolic bottlenecks that formally bound the generative hypothesis space.

\begin{figure*}[t]
\centering
\includegraphics[width=\textwidth]{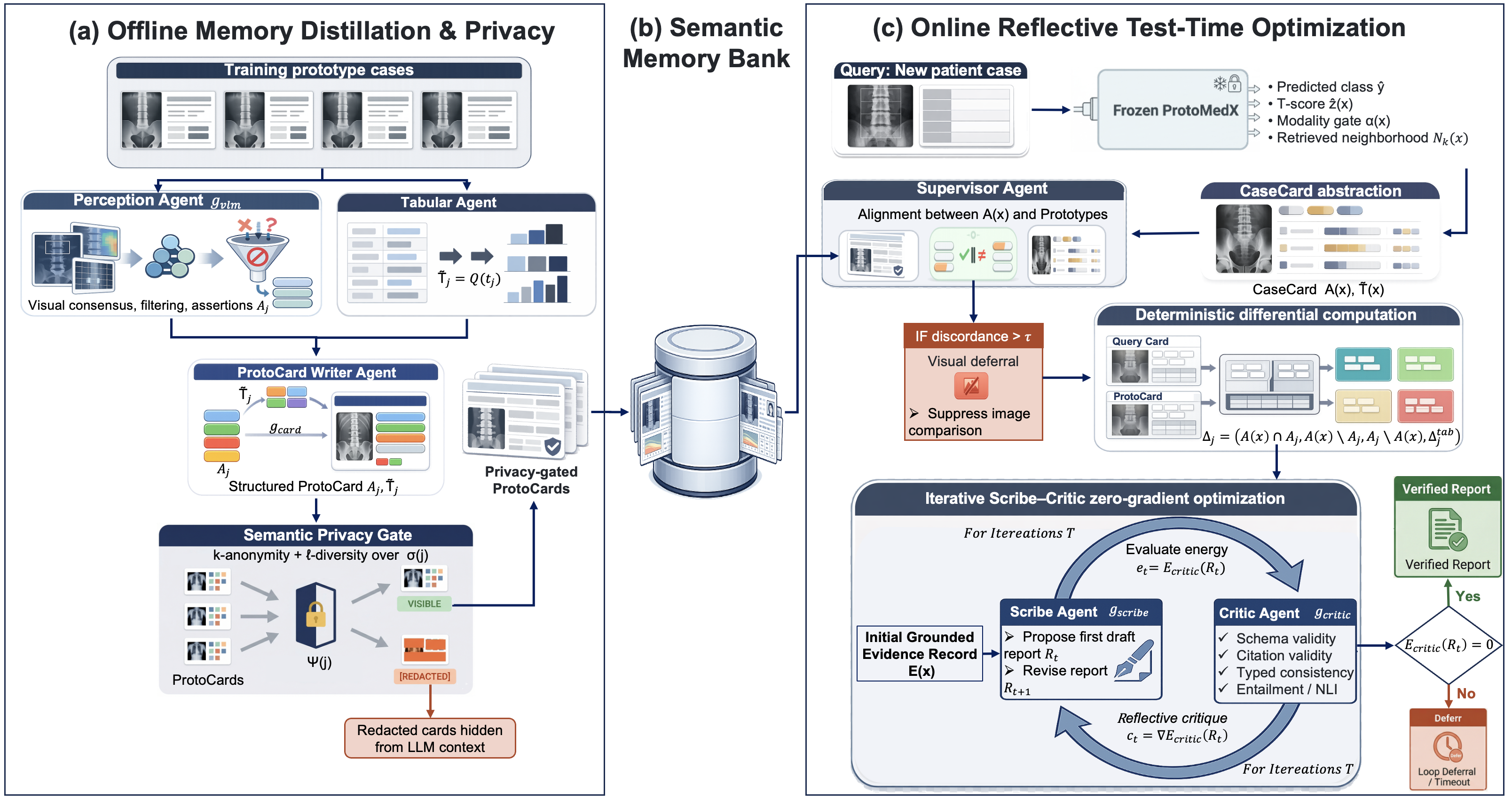}
\caption{Detailed architecture of \protomedagent. (a) \textbf{Offline Memory Distillation \& Privacy} abstracts raw images and tabular records into a discrete semantic memory bank governed by a $k$-anonymity and $\ell$-diversity privacy gate. (b) The intermediate \textbf{Semantic Memory Bank} securely stores verifiable ProtoCards. (c) \textbf{Online Reflective Test-Time Optimization} maps a new patient query into a CaseCard, computes exact set-theoretic differentials ($\Delta_j$), and applies a zero-gradient Scribe-Critic optimization loop to decode the frozen \protomedx backbone into a verifiable report.}
\label{fig:method}
\end{figure*}

Beyond the challenge of hallucination and other vulnerabilities~\cite{LopezPellicer2025ASDefense}, translating retrieved training exemplars into natural language exposes compound systems to severe privacy vulnerabilities. Surfacing raw exemplars during retrieval creates direct vectors for membership and attribute inference attacks~\cite{shokri2017membership, aia}. While Differential Privacy (DP) offers robust theoretical bounds via noise injection~\cite{papernot2018pate, gursoy2017dpnn}, it often catastrophically degrades the exact clinical fidelity required in dense medical imagery. Alternatively, classical $k$-anonymity~\cite{sweeney2002kanon} and its semantic extensions, such as $\ell$-diversity~\cite{machanavajjhala2005ldiversity} and $t$-closeness~\cite{li2007tcloseness}, formalize disclosure risk without injecting continuous noise. By conceptualizing these classical privacy mechanisms as a non-differentiable Markov bottleneck prior to LLM context aggregation, it becomes possible to mathematically secure generated text artifacts against exact linkage and attribute inference, thereby satisfying stringent medical data regulations~\cite{gdpr2016, hipaa164514} without modifying the underlying training dynamics. Beyond privacy law, this design is also well aligned with emerging EU digital regulation. The EU AI Act foregrounds risk management, technical documentation, logging, deployer-facing transparency, human oversight, and lifecycle accuracy/robustness for high-risk AI systems, while the EU Data Act reinforces the value of controlled data access, portability, and interoperable data-processing infrastructures. Together, these trends further motivate privacy-aware, modular, and auditable semantic abstractions for clinical AI deployment~\cite{eu_ai_act_2024,eu_data_act_2023}.

\section{ProtoMedAgent}
\label{sec:method}

\protomedagent conceptualizes clinical reporting as an evidence-grounded translation problem under explicit test-time constraints, as illustrated in Figure \ref{fig:method}. By preserving a frozen predictive backbone and introducing a neuro-symbolic bottleneck, language generation is restricted to operate as an iterative, deterministic interpreter rather than an unconstrained confabulator. Concretely, we decouple prediction from reporting by constraining the generative process to operate over a verified, discrete semantic state derived from prototype evidence.

\subsection{Problem Setting}

Let a patient query be $x=(I,\mathbf{t})$, comprising a high-dimensional visual input $I$ and structured tabular features $\mathbf{t}$. We utilize a frozen prototype network, \protomedx~\cite{Pellicer_2025_ICCV}, which maps $x$ to a predicted class $\hat y$, an auxiliary severity scalar $\hat z(x)$, a modality attention gate $\alpha(x)\in[0,1]$, and a retrieved prototype neighborhood:
$$ \mathcal{N}_k(x)=\{(j,w_j,s_j)\}_{j=1}^{k} $$
where $j$ is the prototype index, $w_j$ the voting weight, and $s_j$ the continuous retrieval similarity.

Our goal is to generate a natural language report $R$ that maximizes clinical fluency while strictly adhering to the evidence encoded in $\mathcal{N}_k(x)$. To achieve this without gradient updates to the generative model, we project $x$ and $\mathcal{N}_k(x)$ into a shared, discrete semantic space of structured clinical assertions, computing exact morphological and clinical differentials that define the admissible semantic state prior to LLM-in-the-loop generation.

\subsection{Offline Memory Distillation \& Privacy}
\label{sec:distill}

Directly exposing raw training images ($I_j$) to an LLM invites sycophantic hallucinations and introduces unacceptable privacy attack vectors. We distill training prototypes $p_j=(I_j,\mathbf{t}_j,y_j)$ into a discrete memory bank of \emph{ProtoCards}, $C_j$.

\paragraph{Perceptual and Tabular Abstraction.}
A radiology Vision-Language Model (VLM), $g_{\text{vlm}}$, serves as a Perception Agent. To extract robust features, $g_{\text{vlm}}$ evaluates $I_j$ under $R$ complementary prompting views. We apply embedding-based clustering and contradiction filtering to extract a conservative set of visual assertions $A_j$. Concurrently, a deterministic Tabular Agent projects continuous patient variables to quantized bins $\tilde T_j = Q(\mathbf{t}_j)$. A schema-constrained generator $g_{\text{card}}$ formats the tuple $(A_j,\tilde T_j)$ into a structured ProtoCard $C_j$.

\paragraph{Semantic Privacy Gate.}
\label{sec:privacy}
We operationalize a semantic privacy gate acting as a non-differentiable regularizer over the memory bank. We enforce a pragmatic $k$-anonymity and $\ell$-diversity rule over a release signature $\sigma(j)$, derived from quantized clinical fields and semantic buckets of the consensus assertions. The visible export surface is:
\begin{equation}
\Psi(j)=
\begin{cases}
C_j, & \begin{aligned}
        &\text{if } \mathrm{cnt}(\sigma(j)) \ge k \\
        &\text{and } \mathrm{ldiv}(\sigma(j)) \ge \ell,
      \end{aligned} \\
\texttt{[REDACTED]}, & \text{otherwise.}
\end{cases}
\end{equation}
Redacted prototypes contribute to the diagnostic forward pass but are completely obfuscated from the LLM's test-time reasoning context.

\subsection{Online Reflective Test-Time Optimization}

During inference, the query $x$ is similarly abstracted into a CaseCard containing visual assertions $A(x)$ and quantized data $\tilde T(x)$. 

\paragraph{Deterministic Differentials and Visual Deferral.}
To prevent the LLM from hallucinating correlations, a Supervisor agent evaluates the alignment between $A(x)$ and the retrieved prototype signatures. If discordance exceeds a threshold $\tau$, the framework triggers \emph{visual deferral}, suppressing image-based similarities to avoid forcing the model to rationalize contradictions. 

For visible prototypes $\bar C_j=\Psi(j)$, we bypass heuristic LLM-driven reasoning by computing an exact set-theoretic differential:
$$ \Delta_j = \big(A(x)\cap A_j,\; A(x)\setminus A_j,\; A_j\setminus A(x),\; \Delta^{\text{tab}}_j\big) $$
where $\Delta^{\text{tab}}_j$ represents the deterministic mismatch in clinical factors. 

\paragraph{Zero-Gradient LLM-in-the-Loop Search.}
We frame report generation as an iterative, zero-gradient search over the discrete state space of natural language reports $\mathcal{R}$. The Critic agent $g_{\text{critic}}$ parameterizes an energy function $E_{\text{critic}}: \mathcal{R} \to \mathbb{R}_{\ge 0}$ defined by a multi-objective constraint:
$$ E_{\text{critic}}(R_t) = \lambda_1 \mathcal{L}_{\text{schema}} + \lambda_2 \mathcal{L}_{\text{cite}} + \lambda_3 \mathcal{L}_{\text{type}} + \lambda_4 \mathcal{L}_{\text{NLI}} $$
where the loss terms penalize JSON invalidity, citation hallucination against the grounded state $E(x)$, typed-value inconsistency, and logical entailment failures. The Scribe agent $g_{\text{scribe}}$ acts as a conditional proposal distribution $P_{\text{scribe}}(R_{t+1} \mid E(x), \nabla E_{\text{critic}}(R_t))$, where the gradient $\nabla E_{\text{critic}}$ is instantiated as structured natural language reflections. This process is formalized in Algorithm \ref{alg:optimization}.

\begin{algorithm}[h]
\caption{LLM-in-the-Loop Reflective Optimization}
\label{alg:optimization}
\begin{algorithmic}[1]
\Require Grounded state $E(x)$, max iterations $T$, energy threshold $\epsilon=0$
\State $R_0 \sim P_{\text{scribe}}(\cdot \mid E(x))$
\For{$t = 0$ to $T-1$}
    \State Evaluate energy $e_t = E_{\text{critic}}(R_t)$
    \If{$e_t \le \epsilon$}
        \State \Return $R_t$ \Comment{Convergence to faithful state}
    \EndIf
    \State Generate reflective critique $c_t \gets \nabla E_{\text{critic}}(R_t)$
    \State Sample new report $R_{t+1} \sim P_{\text{scribe}}(\cdot \mid E(x), c_t)$
\EndFor
\State \Return \textsc{Defer} \Comment{Failed to satisfy neuro-symbolic constraints}
\end{algorithmic}
\end{algorithm}

\subsection{Theoretical Guarantees of the Framework}
\label{sec:theory}
By restructuring multimodal reporting as an energy-based optimization problem over a neuro-symbolic bottleneck, \protomedagent provides two distinct operational guarantees regarding generative faithfulness and privacy exposure. Formal proofs for both theorems are provided in Supplementary Section B.

\textbf{Theorem 1 (Bounded Generative Faithfulness):} \emph{Let $\mathcal{R}$ be the hypothesis space of all possible natural language reports, and let the deterministic critic parameterize a barrier energy function $E_{\text{critic}}: \mathcal{R} \to \{0, \infty\}$ indicating an admissible constraint set. If the optimization converges such that $E_{\text{critic}}(R^*) = 0$, the set of atomic semantic claims within the accepted report $R^*$ is strictly bounded by the deterministic differential $\Delta_j$, mathematically precluding ungrounded generative confabulation.} 

This theorem establishes that while the underlying Vision-Language Model (VLM) dictates the absolute visual truth of the system, the generative LLM is restricted from hallucinating structural or tabular comparisons outside the pre-computed intersection $(A(x)\cap A_j)$ and relative complements $(A(x)\setminus A_j, A_j\setminus A(x))$. The generative hypothesis space is thereby safely bounded to strict interpretation of the symbolic state rather than open-ended invention.

\textbf{Theorem 2 (Information-Theoretic Privacy Bound):} \emph{Assuming the export function $\Psi(j)$ enforces $k$-anonymity and strict $\ell$-diversity, and an adversary $\mathcal{A}$ possesses a prior that matches the empirical distribution of the equivalence class (i.e., no auxiliary background knowledge), the maximum posterior probability of $\mathcal{A}$ successfully executing an exact record linkage attack from the generated report $R^*$ is upper-bounded by $1/k$, and the probability of targeted attribute inference is upper-bounded by $1/\ell$.}

Because the semantic privacy gate acts as a non-differentiable Markov bottleneck prior to test-time aggregation, the generative model operates exclusively on a sanitized context space. By the conditional independence established by the Markov property, the LLM cannot reconstruct case-specific signatures destroyed by $\Psi(j)$, neutralizing artifact-level inversion attacks regardless of the LLM's learned semantic priors.
\section{Experiments}
\label{sec:experiments}

\subsection{Task and dataset}
We study three-class bone-health classification (\textsc{Normal}, \textsc{Osteopenia}, \textsc{Osteoporosis}) from lumbar-spine DEXA scans and structured clinical risk factors.
We follow the dataset and protocol of \protomedx~\cite{Pellicer_2025_ICCV}: 4,160 de-identified NHS patients with 11 FRAX\textsuperscript{\textregistered}-aligned variables, using the published preprocessing pipeline and an 80/20 train--test split with an internal 10\% validation split.

\subsection{Frozen encoder \&  memory}

All experiments use one frozen \protomedx checkpoint. For every case, the report writer receives the same backbone prediction $\hat y$, auxiliary T-score $\hat z(x)$, modality gate $\alpha(x)$, and retrieved neighborhood $\mathcal N_k(x)$. The backbone is never fine-tuned inside the reporting experiments, so any differences in report quality arise from the evidence-to-language interface rather than from re-training the predictor.

\subsection{Backbone Agent choices and baselines}
For the Perception Agent, we use openly available medical VLMs (RadFM~\cite{wu2023radfm} and LLaVA-Med~\cite{li2023llavamed}) to extract atomic image assertions under multiple prompts, followed by the consensus step described in Sec.~\ref{sec:distill}. For ProtoCard writing and report generation, we use medical instruction models such as BioMistral~\cite{labrak2024biomistral} and Meditron~\cite{chen2023meditron} under structured decoding \cite{willard2023outlines,dong2024xgrammar}.

We compare against two reporting baselines that operate on the same frozen \protomedx outputs. \textbf{Standard RAG} conditions an LLM on the retrieved evidence and directly generates a report. \textbf{LLM (LoRA)} uses a domain-adapted schema-filling model~\cite{hu2022lora}, but still relies on the model to infer case--prototype agreement implicitly. \textbf{\protomedagent} adds privacy-gated ProtoCards, deterministic comparison objects, supervisor-controlled visual deferral, and a final Scribe--Critic verification loop.

\subsection{Evaluation protocol}
\protomedagent is accuracy-neutral, so we evaluate the reporting layer rather than predictive accuracy. Our primary metric is \textbf{Comparison Set Faithfulness} (CSF): for each query--prototype pair, we derive deterministic reference sets of shared and differing evidence items from the comparison object $\Delta_j$, extract the items verbalized in the final report, and compute CSF precision, recall, F1, and weighted accuracy. These metrics test whether a report preserves the case--prototype comparison actually available to the system.

We also evaluate the\textbf{ privacy--utility} behavior of the released evidence surface. First, we sweep the release gate over different $(k,\ell)$ values and report similarity-weighted evidence utility, visible-card rate, redaction rate, and top-1 linkage success rate (\textbf{Link.}) on the ProtoCard surface. Second, at the selected operating point, we evaluate three artifact-level\textbf{ disclosure attacks} on the released outputs: Membership-Inference Accuracy (\textbf{MIA}), which tests whether an attacker can determine whether the underlying case belongs to the protected source corpus \cite{shokri2017membership}; Attribute-Inference Accuracy (\textbf{AIA}), which tests whether sensitive source attributes can be inferred from the released artifact, potentially with auxiliary information \cite{aia}; and top-1 Linkage success rate (\textbf{Link.}), which tests whether sparse released cues suffice to re-identify the correct source record \cite{Ko2026InferenceDrivenLinkage}. These attacks are directly relevant because ProtoMedAgent's privacy gate and non-differentiable Markov bottleneck are designed to suppress residual membership leakage, sensitive-attribute disclosure, and exact re-identification risk. 

Formal definitions for each of our evaluation metrics are provided in Supplementary material section A.

\section{Results}
\label{sec:results}

Because \protomedagent operates strictly as a test-time reporting layer over a frozen \protomedx backbone, it is designed to be accuracy-neutral with respect to the underlying predictive task. Our evaluation therefore isolates the efficacy of the agentic translation layer. Specifically, we investigate whether formulating report generation as an iterative optimization over a discrete neuro-symbolic bottleneck can enforce strict comparison faithfulness and bound privacy exposure, without sacrificing clinical fluency. Accordingly, all reported improvements reflect gains in reporting fidelity, safety, and privacy rather than predictive performance.

\subsection{Safety and Comparison Faithfulness}

To evaluate the fidelity of the generated reports against the retrieved visual evidence (answering RQ2), we introduce the Comparison Set Faithfulness (CSF) metric. CSF isolates the explicit matches and differences asserted in the generated text and evaluates them against the deterministic, set-theoretic differentials ($\Delta_j$) computed between the query and the retrieved ProtoCards, providing a direct measure of evidence-grounded comparison fidelity. 

As reported in Table~\ref{tabsaf}, \protomedagent achieves state-of-the-art performance for grounded clinical reporting. By decoupling the diagnostic forward pass from evidence documentation, our framework restricts the Scribe agent to narrating only those evidence-supported partitions backed by exact citations. Furthermore, the iterative Critic-repair loop promotes item-level consistency: any hallucinated correlation that cannot be traced to the admissible semantic state is dynamically corrected or removed. This behavior is consistent with Theorem~1, which bounds the generative space to evidence-supported differentials. 

This mechanism directly addresses the vulnerability of standard generative approaches. Baselines such as Standard RAG \cite{lewis2020rag} and domain-adapted LoRA \cite{hu2022lora} rely on implicit, unconstrained reasoning over retrieved contexts. Without explicit neuro-symbolic constraints, these models are prone to retrieval sycophancy, inventing post-hoc rationalizations to align with the backbone's prediction. This failure mode is empirically confirmed by their degradation in CSF-Precision (46.2\% and 63.1\%, respectively) compared to the bounded output of \protomedagent (91.2\%).

\begin{table}[t]
\centering
\caption{Safety and Faithfulness Study on the test cohort. By strictly bounding the generative hypothesis space to pre-computed deterministic differentials, \protomedagent fundamentally mitigates multimodal hallucination and unsupported claims.}
\label{tabsaf}
\setlength{\tabcolsep}{4pt}
\renewcommand{\arraystretch}{1.05}
\resizebox{\columnwidth}{!}{
\begin{tabular}{lcccc}
\toprule
\textbf{Method} & \textbf{CSF-P $\uparrow$} & \textbf{CSF-R $\uparrow$} & \textbf{CSF-F1 $\uparrow$} & \textbf{CSF-WA $\uparrow$} \\
\midrule
Standard RAG & 46.2 & 41.5 & 43.7 & 58.6 \\
LLM (LoRA)   & 63.1 & 58.9 & 60.9 & 70.2 \\
\textbf{Ours} & \textbf{91.2} & \textbf{78.0} & \textbf{84.1} & \textbf{92.7} \\
\bottomrule
\end{tabular}
}
\end{table}

\begin{figure}[t]
\centering
\includegraphics[width=\linewidth]{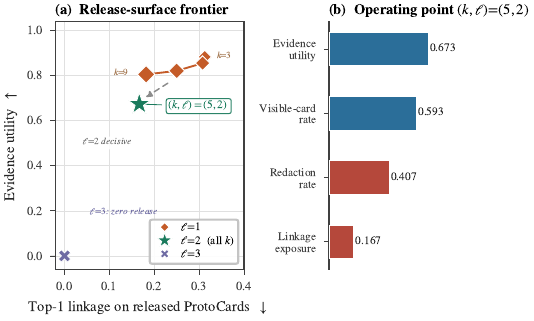}
\caption{\textbf{Privacy-utility frontier of the exportable prototype evidence surface.} Sweeping semantic $k$-anonymity ($k\!\in\!\{3,5,7,9\}$) and $\ell$-diversity ($\ell\!\in\!\{1,2,3\}$). \textbf{(a)} Under $\ell\!=\!1$ (orange), increasing $k$ provides a smooth trade-off between utility and exposure. Imposing $\ell\!=\!2$ (green) triggers a phase transition, collapsing all $k$ values to a single bounded regime, indicating diversity binds before $k$-anonymity. \textbf{(b)} The selected operating point $(k,\ell)\!=\!(5,2)$ retains robust evidence utility (0.673) while suppressing linkage exposure to 0.167.}
\label{fig:privacy_frontier}
\end{figure}

\begin{figure*}[t]
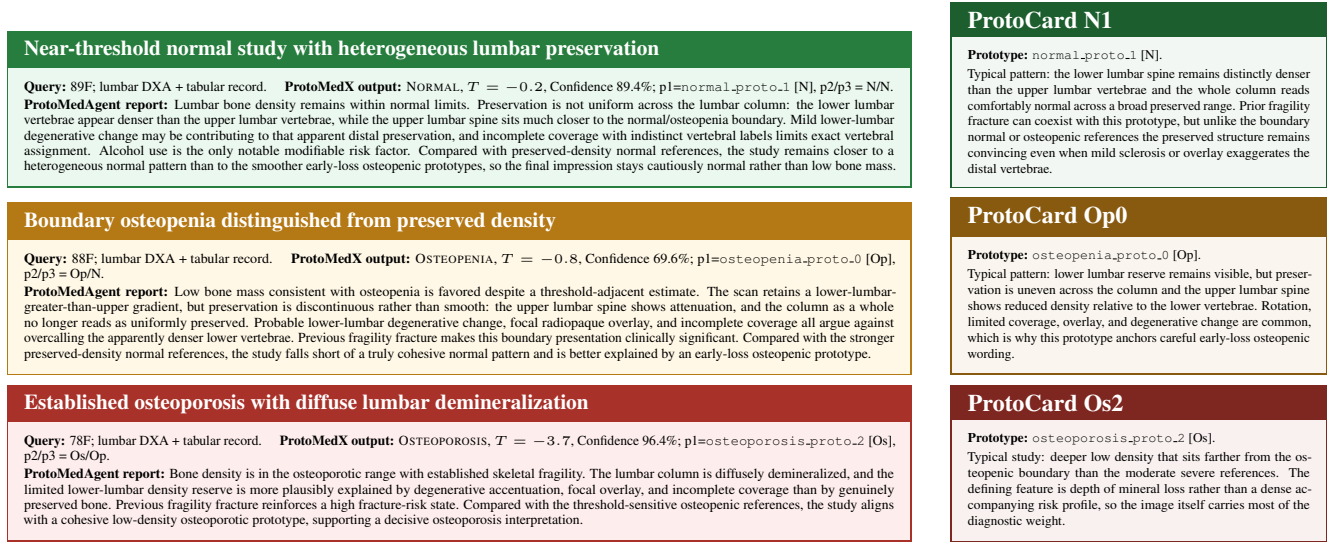

\centering
\fontsize{5.2}{5.8}\selectfont
\setlength{\tabcolsep}{0pt}
\renewcommand{\arraystretch}{1.0}

\begin{tabular*}{\textwidth}{@{\extracolsep{\fill}}p{0.685\textwidth}p{0.285\textwidth}@{}}
\parbox[t]{\linewidth}{%
\begin{tcolorbox}[
  enhanced,
  sharp corners=all,
  boxrule=0.6pt,
  colback=clrBgNormal,
  colframe=clrNormal,
  colbacktitle=clrNormal,
  coltitle=white,
  fonttitle=\bfseries\footnotesize,
  title={Near-threshold normal study with heterogeneous lumbar preservation},
  width=\linewidth,
  breakable=false,
  left=3pt,right=3pt,top=2pt,bottom=2pt,
  nobeforeafter]
\textbf{Query:} 89F; lumbar DXA + tabular record. \quad
\textbf{ProtoMedX output:} \textsc{Normal}, $T=-0.2$, Confidence 89.4\%; p1=\texttt{normal\_proto\_1} [N], p2/p3 = N/N.
\par\vspace{1pt}
\textbf{ProtoMedAgent report:} Lumbar bone density remains within normal limits. Preservation is not uniform across the lumbar column: the lower lumbar vertebrae appear denser than the upper lumbar vertebrae, while the upper lumbar spine sits much closer to the normal/osteopenia boundary. Mild lower-lumbar degenerative change may be contributing to that apparent distal preservation, and incomplete coverage with indistinct vertebral labels limits exact vertebral assignment. Alcohol use is the only notable modifiable risk factor. Compared with preserved-density normal references, the study remains closer to a heterogeneous normal pattern than to the smoother early-loss osteopenic prototypes, so the final impression stays cautiously normal rather than low bone mass.
\end{tcolorbox}} &
\parbox[t]{\linewidth}{%
\begin{tcolorbox}[
  enhanced,
  sharp corners,
  boxrule=0.6pt,
  colback=clrNormal!6,
  colframe=clrNormal!75!black,
  title={\small\bfseries ProtoCard N1},
  width=\linewidth,
  breakable=false,
  left=3pt,right=3pt,top=2pt,bottom=2pt,
  nobeforeafter]
\tiny
\textbf{Prototype:} \texttt{normal\_proto\_1} [N].
\par\vspace{1pt}
Typical pattern: the lower lumbar spine remains distinctly denser than the upper lumbar vertebrae and the whole column reads comfortably normal across a broad preserved range. Prior fragility fracture can coexist with this prototype, but unlike the boundary normal or osteopenic references the preserved structure remains convincing even when mild sclerosis or overlay exaggerates the distal vertebrae.
\end{tcolorbox}}
\\[2pt]
\parbox[t]{\linewidth}{%
\begin{tcolorbox}[
  enhanced,
  sharp corners=all,
  boxrule=0.6pt,
  colback=clrBgOsteopenia,
  colframe=clrOsteopenia,
  colbacktitle=clrOsteopenia,
  coltitle=white,
  fonttitle=\bfseries\footnotesize,
  title={Boundary osteopenia distinguished from preserved density},
  width=\linewidth,
  breakable=false,
  left=3pt,right=3pt,top=2pt,bottom=2pt,
  nobeforeafter]
\textbf{Query:} 88F; lumbar DXA + tabular record. \quad
\textbf{ProtoMedX output:} \textsc{Osteopenia}, $T=-0.8$, Confidence 69.6\%; p1=\texttt{osteopenia\_proto\_0} [Op], p2/p3 = Op/N.
\par\vspace{1pt}
\textbf{ProtoMedAgent report:} Low bone mass consistent with osteopenia is favored despite a threshold-adjacent estimate. The scan retains a lower-lumbar-greater-than-upper gradient, but preservation is discontinuous rather than smooth: the upper lumbar spine shows attenuation, and the column as a whole no longer reads as uniformly preserved. Probable lower-lumbar degenerative change, focal radiopaque overlay, and incomplete coverage all argue against overcalling the apparently denser lower vertebrae. Previous fragility fracture makes this boundary presentation clinically significant. Compared with the stronger preserved-density normal references, the study falls short of a truly cohesive normal pattern and is better explained by an early-loss osteopenic prototype.
\end{tcolorbox}} &
\parbox[t]{\linewidth}{%
\begin{tcolorbox}[
  enhanced,
  sharp corners,
  boxrule=0.6pt,
  colback=clrOsteopenia!7,
  colframe=clrOsteopenia!75!black,
  title={\small\bfseries ProtoCard Op0},
  width=\linewidth,
  breakable=false,
  left=3pt,right=3pt,top=2pt,bottom=2pt,
  nobeforeafter]
\tiny
\textbf{Prototype:} \texttt{osteopenia\_proto\_0} [Op].
\par\vspace{1pt}
Typical pattern: lower lumbar reserve remains visible, but preservation is uneven across the column and the upper lumbar spine shows reduced density relative to the lower vertebrae. Rotation, limited coverage, overlay, and degenerative change are common, which is why this prototype anchors careful early-loss osteopenic wording.
\end{tcolorbox}}
\\[2pt]
\parbox[t]{\linewidth}{%
\begin{tcolorbox}[
  enhanced,
  sharp corners=all,
  boxrule=0.6pt,
  colback=clrBgOsteoporosis,
  colframe=clrOsteoporosis,
  colbacktitle=clrOsteoporosis,
  coltitle=white,
  fonttitle=\bfseries\footnotesize,
  title={Established osteoporosis with diffuse lumbar demineralization},
  width=\linewidth,
  breakable=false,
  left=3pt,right=3pt,top=2pt,bottom=2pt,
  nobeforeafter]
\textbf{Query:} 78F; lumbar DXA + tabular record. \quad
\textbf{ProtoMedX output:} \textsc{Osteoporosis}, $T=-3.7$, Confidence 96.4\%; p1=\texttt{osteoporosis\_proto\_2} [Os], p2/p3 = Os/Op.
\par\vspace{1pt}
\textbf{ProtoMedAgent report:} Bone density is in the osteoporotic range with established skeletal fragility. The lumbar column is diffusely demineralized, and the limited lower-lumbar density reserve is more plausibly explained by degenerative accentuation, focal overlay, and incomplete coverage than by genuinely preserved bone. Previous fragility fracture reinforces a high fracture-risk state. Compared with the threshold-sensitive osteopenic references, the study aligns with a cohesive low-density osteoporotic prototype, supporting a decisive osteoporosis interpretation.
\end{tcolorbox}} &
\parbox[t]{\linewidth}{%
\begin{tcolorbox}[
  enhanced,
  sharp corners,
  boxrule=0.6pt,
  colback=clrOsteoporosis!7,
  colframe=clrOsteoporosis!75!black,
  title={\small\bfseries ProtoCard Os2},
  width=\linewidth,
  breakable=false,
  left=3pt,right=3pt,top=2pt,bottom=2pt,
  nobeforeafter]
\tiny
\textbf{Prototype:} \texttt{osteoporosis\_proto\_2} [Os].
\par\vspace{1pt}
Typical study: deeper low density that sits farther from the osteopenic boundary than the moderate severe references. The defining feature is depth of mineral loss rather than a dense accompanying risk profile, so the image itself carries most of the diagnostic weight.
\end{tcolorbox}}
\end{tabular*}

\caption{\textbf{Final ProtoMedAgent qualitative outputs.} Each row pairs a final ProtoMedAgent report at left with the displayed same-class ProtoCard reference at right. The header line shows the query summary and compact ProtoMedX output; $p1$ is the ProtoCard shown beside the report, while $p2/p3$ summarize the remaining retrieved neighborhood. Here N = \textsc{Normal}, Op = \textsc{Osteopenia}, and Os = \textsc{Osteoporosis}. The three examples show a cautious near-threshold normal interpretation, a boundary osteopenic case clarified by an early-loss osteopenic reference, and a decisively osteoporotic study anchored by deep low-density prototype evidence. Supplementary Sec.~S4 expands the report gallery to routine, restricted-evidence, and harder boundary cases, and Supplementary Sec.~S5 provides the full prototype atlas.}
\label{fig:qualitative_main}
\end{figure*}

\subsection{Privacy-Utility Dynamics \& Disclosure Risk}

We evaluate the semantic privacy gate (answering RQ1) through a dual-lens approach: (i) a \emph{release-surface analysis} mapping the $(k,\ell)$ configuration space to characterize the privacy-utility frontier, and (ii) rigorous \emph{artifact-level attack probes} measuring residual disclosure risk on the final generated text.

A key empirical finding illustrated in Fig.~\ref{fig:privacy_frontier}(a) is the phase transition governed by $\ell$-diversity. Under $\ell\!=\!1$, scaling $k$-anonymity yields a smooth, proportional decay in both evidence utility and linkage exposure. However, enforcing $\ell\!=\!2$ causes all tested $k$ values to collapse into an identical, highly constrained operating regime (utility\,=\,0.673, linkage\,=\,0.167). This indicates that the diversity of the underlying clinical attributes becomes the dominant constraint prior to cohort size ($k$-anonymity), effectively saturating the admissible equivalence classes.

This behavior reveals an important structural property of the semantic release surface: privacy constraints do not combine linearly. Instead, $\ell$-diversity introduces a combinatorial constraint over attribute distributions, which sharply reduces the feasible support of the exportable memory. As a result, once diversity is enforced, increasing $k$ yields diminishing returns, explaining the observed collapse across all $k$ values. Pushing the constraint further to $\ell\!=\!3$ uniformly eliminates valid equivalence classes, reducing the visible surface to zero.

At the selected operating point of $(k,\ell)\!=\!(5,2)$, the gate retains 32 of 54 candidate ProtoCards (59.3\% visibility) and deliberately enforces a 40.7\% redaction rate. Importantly, this redaction is not random but structured: removed prototypes correspond to semantically rare or uniquely identifying clinical configurations. This confirms our theoretical claim that privacy-aware prototype reporting should be formulated as an explicit operating-point optimization over a semantic release surface, rather than treating raw exemplar exposure as a binary heuristic.

\begin{table}[t]
\centering
\caption{Artifact-level disclosure risk on exported evidence. Metrics include Membership-Inference Accuracy (MIA), Attribute-Inference Accuracy (AIA), and top-1 Linkage success rate (Link.). \protomedagent systematically lowers empirical attack vulnerabilities by sanitizing the generative context window.}
\label{tabpri}
\setlength{\tabcolsep}{5pt}
\renewcommand{\arraystretch}{1.05}
\begin{tabular}{l c c c}
\toprule
\textbf{Method} & \textbf{MIA $\downarrow$} & \textbf{AIA $\downarrow$} & \textbf{Link.\ $\downarrow$} \\
\midrule
Standard RAG & 51.2 & 34.2 & 39.3 \\
LLM (LoRA)   & 49.7 & 33.0 & 38.1 \\
\textbf{Ours} & \textbf{41.4} & \textbf{25.8} & \textbf{34.5} \\
\bottomrule
\end{tabular}
\end{table}

Table~\ref{tabpri} measures the residual disclosure risk remaining in the final text artifacts. Because \protomedagent enforces generation through a non-differentiable Markov bottleneck, the LLM is effectively prevented from reconstructing case-specific signatures. This constraint reduces the effective attack surface by restricting generation to aggregated, equivalence-class-level information rather than instance-level features.

This theoretical protection translates into robust empirical defense, yielding the lowest risk profiles across all metrics, notably reducing MIA by an absolute 9.8\% and AIA by 8.4\% relative to Standard RAG. Importantly, these reductions are achieved without modifying the underlying generative model, indicating that controlling the information interface—rather than the model parameters—is sufficient to significantly mitigate disclosure risk.

\subsection{Qualitative Results}

While quantitative metrics confirm strict neuro-symbolic adherence, Figure~\ref{fig:qualitative_main} demonstrates that this bottleneck does not materially degrade clinical fluency. The examples illustrate how the LLM maps discrete ProtoCard references into cohesive, highly contextual natural language.

Crucially, the qualitative variations across these reports are clinically substantive rather than stochastic artifacts of the language model. For instance, in the threshold-adjacent osteopenic study, the narrative does not simply echo a generic class label. Instead, it grounds its early mineral loss interpretation by explicitly contrasting the scan against stronger preserved-density normal prototypes, carefully acknowledging the residual lower-lumbar reserve. Conversely, the osteoporotic report synthesizes diffuse demineralization, historical fragility fractures, and alignment with a severe low-density reference prototype to anchor a coherent structural degradation narrative.

Beyond descriptive fidelity, these examples highlight three important behavioral properties of \protomedagent. First, the generated reports exhibit \emph{evidence-calibrated reasoning}: claims are modulated by the strength and consistency of retrieved prototypes, with boundary cases producing more nuanced, comparative language rather than overconfident categorical statements. Second, the framework demonstrates \emph{counterfactual sensitivity}: small shifts in the retrieved prototype neighborhood lead to controlled and interpretable changes in the generated narrative, indicating that the LLM is responding to structured evidence rather than latent priors. Third, the system enforces \emph{faithful abstraction}: while the wording remains fluent and expressive, all semantic content can be traced back to admissible set-theoretic differentials, preventing unsupported clinical assertions.

Importantly, we also observe that when evidence is partially redacted by the privacy gate, the model adapts by producing more conservative and less specific language, rather than hallucinating missing details. This behavior suggests that the neuro-symbolic bottleneck induces a form of implicit uncertainty awareness in the generative process.

By displaying the cited ProtoCards alongside the generated text, \protomedagent renders the LLM's wording fully auditable in concrete visual and tabular terms. Notably, all narrative variations remain grounded in the admissible semantic state, demonstrating that expressivity is preserved without sacrificing faithfulness. Supplementary section C expands this qualitative analysis to include complex boundary cases and restricted-evidence scenarios, while the complete prototype atlas is provided in Supplementary material section D.
\section{Discussion and Conclusion}
\label{sec:discussion}

\protomedagent formalizes \textbf{multimodal prototype retrieval} as a constrained, test-time optimization problem, effectively bridging the semantic gap between continuous representations and discrete clinical documentation. By executing a strict Scribe$\leftrightarrow$Critic loop over a privacy-gated semantic memory and a frozen \protomedx backbone, we reposition the LLM as a verifiable interpretability layer. This approach achieves \textbf{state-of-the-art Comparison Set Faithfulness} (91.2\% CSF-Precision), proving that generative faithfulness can be structurally enforced without costly model parameter modifications. Furthermore, framing\textbf{ privacy-aware reporting} as an explicit operating-point optimization reveals a binding $\ell$-diversity phase transition. This non-differentiable Markov bottleneck severely restricts case-specific information flow, delivering robust empirical defense. Including a \textbf{9.8\% absolute reduction in Membership Inference attacks} without altering the underlying weights.

Despite these systemic guarantees, the framework remains bounded by its upstream Perception Agent, which is vulnerable to correlated visual blind spots. Additionally, our heuristic Discordance Sentinel introduces an inherent tension between generative expressivity and conservative visual deferral. From a privacy perspective, semantic $k$-anonymity and $\ell$-diversity lack the worst-case theoretical guarantees of Differential Privacy, leaving residual exposure to adversaries with out-of-schema auxiliary knowledge. Crucially, while our neuro-symbolic bottleneck mathematically precludes hallucinated structural claims, it strictly bounds the \emph{space of admissible explanations} without guaranteeing the absolute \emph{truthfulness} of the underlying backbone's perception.

Nevertheless, as a non-diagnostic, assistive auditing tool that supports explicit human deferral and continuous ``patchability'' of prototype memories, \protomedagent aligns directly with the stringent technical transparency and human oversight mandates of the \textbf{EU AI Act and EU Data Act}. 

To fully research this clinical utility, a critical next step is \textbf{conducting controlled user studies} to rigorously validate our qualitative findings and assess the framework's true impact on decision-making workflows. Methodologically, future research should prioritize integrating calibrated uncertainty into the perception layer, formalizing citation semantics via schema-versioned references (e.g., JSONPath), and exploring hybrid DP-symbolic architectures to extend formal privacy bounds to continuous prototype systems.
\section*{Acknowledgment}
This study is approved by the North West- Preston Research Ethics Committee (REC ref 21/NW/0309).
This work is supported by ELSA – European Lighthouse on Secure and Safe AI funded by the European Union under grant agreement No. 101070617.
{\small
\bibliographystyle{ieeenat_fullname}
\bibliography{main}
}

\clearpage
\onecolumn 

\appendix

\setcounter{page}{1}
\setcounter{section}{0}
\setcounter{equation}{0}
\setcounter{figure}{0}
\setcounter{table}{0}

\renewcommand{\thesection}{S\arabic{section}}
\renewcommand{\theequation}{S\arabic{equation}}
\renewcommand{\thefigure}{S\arabic{figure}}
\renewcommand{\thetable}{S\arabic{table}}

\begin{center}
    \Large \textbf{\protomedagent: Supplementary Material} \\ \vspace{0.8cm}
\end{center}

\section{Evaluation Metrics}
\label{sec:metrics}

This section provides a detailed definition of the evaluation metrics used in the results of our report.

\paragraph{Comparison Set Faithfulness (CSF).}
CSF measures whether the model's case--prototype comparison claims are consistent with evidence-derived differences~\cite{trace}. For each query case and retrieved prototype, we compute deterministic reference partitions over evidence items $\Delta$, where each item is either a visual assertion ID or a tabular factor-field ID. From the generated report, we extract the predicted partitions $\widehat{\Delta}$ (items claimed as shared, query-only, or prototype-only) using the report schema fields. We then compute:
\begin{equation}
\mathrm{CSF\text{-}Precision}=\frac{|\widehat{\Delta}_{\mathrm{diff}}\cap \Delta_{\mathrm{diff}}|}{|\widehat{\Delta}_{\mathrm{diff}}|}
\end{equation}
\begin{equation}
\mathrm{CSF\text{-}Recall}=\frac{|\widehat{\Delta}_{\mathrm{diff}}\cap \Delta_{\mathrm{diff}}|}{|\Delta_{\mathrm{diff}}|}
\end{equation}
\begin{equation}
\mathrm{CSF\text{-}F1}=\frac{2\,\mathrm{CSF\text{-}Precision}\,\mathrm{CSF\text{-}Recall}}{\mathrm{CSF\text{-}Precision}+\mathrm{CSF\text{-}Recall}},
\end{equation}
where $\Delta_{\mathrm{diff}}=\Delta_{\mathrm{q\text{-}only}}\cup \Delta_{\mathrm{p\text{-}only}}$ and $\widehat{\Delta}_{\mathrm{diff}}=\widehat{\Delta}_{\mathrm{q\text{-}only}}\cup \widehat{\Delta}_{\mathrm{p\text{-}only}}$.

To summarize three-way assignment quality, we compute a weighted accuracy over the partition labels at the item level:
\begin{equation}
\mathrm{CSF\text{-}WA}=\frac{\sum_{c} w_c \cdot \mathrm{Acc}_c}{\sum_{c} w_c}
\end{equation}
\begin{equation}
\mathrm{Acc}_c=\frac{\left|\left\{i: \ell(i)=c \land \hat{\ell}(i)=c\right\}\right|}{\left|\left\{i: \ell(i)=c\right\}\right|},
\end{equation}
where $\ell(i)$ and $\hat{\ell}(i)$ denote the reference and predicted partition labels for item $i$, and $w_c$ are class weights. Precision penalizes unsupported difference claims, recall measures coverage of evidence-backed differences, and CSF-WA summarizes overall three-way partition fidelity.

\paragraph{Privacy evaluation.}
We evaluate privacy at two complementary levels. First, we analyze the \emph{release surface} induced by the export gate by sweeping semantic $k$-anonymity and $\ell$-diversity over the candidate ProtoCard pool. Second, the main paper reports \emph{artifact-level} attacks on the final exported evidence under a shared protocol across methods. In this supplement we therefore focus on the release-surface quantities that support the frontier in the main paper: weighted evidence utility, visible-card rate, redaction rate, and linkage exposure on the released ProtoCard surface.

\begin{table}[h!]
\centering
\small
\caption{Detailed release-surface behavior at the selected operating point $(k,\ell)=(5,2)$.}
\label{tab:s_privacy_operating}
\begin{tabular}{L{0.55\linewidth}cc}
\toprule
\textbf{Measure} & \textbf{Value} & \textbf{Scope} \\
\midrule
Cited evidence utility over held-out reports & 0.369 & 832 reports \\
Similarity-weighted utility on prototype release surface & 0.673 & 54 candidate cards \\
Visible-card rate & 0.593 & 54 candidate cards \\
Redaction rate & 0.407 & 54 candidate cards \\
Top-1 linkage on released ProtoCards & 0.167 & 54 candidate cards \\
\bottomrule
\end{tabular}
\end{table}

\begin{table}[h!]
\centering
\small
\caption{Compressed privacy--utility sweep over semantic $k$-anonymity and $\ell$-diversity. Repeated regimes are merged to emphasize the active control variable.}
\label{tab:s_privacy_sweep_compact}
\begin{tabular}{cccccc}
\toprule
$k$ & $\ell$ & Utility & Visible & Linkage & Interpretation \\
\midrule
3 & 1 & 0.882 & 0.852 & 0.312 & loosest release \\
5 & 1 & 0.854 & 0.796 & 0.308 & smoother trade-off \\
7 & 1 & 0.820 & 0.759 & 0.250 & more restrictive \\
9 & 1 & 0.804 & 0.722 & 0.182 & strongest $k$-only point \\
3,5,7,9 & 2 & 0.673 & 0.593 & 0.167 & identical regime across all tested $k$ \\
3,5,7,9 & 3 & 0.000 & 0.000 & 0.000 & no shareable surface \\
\bottomrule
\end{tabular}
\end{table}

\paragraph{Membership-Inference Accuracy (MIA).}
Let \(o_i\) denote a released artifact associated with source case \(i\), and let
\(m_i \in \{0,1\}\) indicate whether that source case belongs to the private
training/prototype corpus (\(m_i=1\)) or not (\(m_i=0\)).
Given an attack model \(a_{\mathrm{mia}}\), we report
\[
\mathrm{MIA}
=
\frac{1}{N_{\mathrm{mia}}}
\sum_{i=1}^{N_{\mathrm{mia}}}
\mathbf{1}\!\left[a_{\mathrm{mia}}(o_i)=m_i\right].
\]
MIA measures whether the released artifact leaks the membership status of an
individual case. It is relevant here because ProtoMedAgent explicitly seeks to
reduce artifact-level membership leakage from the exported evidence surface and
the final generated outputs.

\paragraph{Attribute-Inference Accuracy (AIA).}
Let \(s_i \in \mathcal{S}\) denote a sensitive attribute of source case \(i\),
and let \(z_i\) denote any allowed auxiliary side information available to the
attacker. If the attack model predicts
\(\hat{s}_i = a_{\mathrm{aia}}(o_i, z_i)\), then we define
\[
\mathrm{AIA}
=
\frac{1}{N_{\mathrm{aia}}}
\sum_{i=1}^{N_{\mathrm{aia}}}
\mathbf{1}\!\left[\hat{s}_i=s_i\right].
\]
AIA measures how accurately an adversary can recover sensitive attributes from
the released artifact, possibly combined with auxiliary information. It is
especially relevant here because the semantic privacy gate is designed to limit
attribute disclosure, which is precisely the threat probed by AIA.

\paragraph{Top-1 Linkage Success Rate (\textnormal{Link.}).}
Let \(o_i\) be a released artifact, let \(\mathcal{C}_i\) be the adversary's
candidate set of identified records, and let \(g(o_i,c)\) be a matching score
between artifact \(o_i\) and candidate record \(c\). The attacker returns the
single best match
\[
\hat{c}_i=\arg\max_{c \in \mathcal{C}_i} g(o_i,c).
\]
Writing \(c_i^\star\) for the true source record of \(o_i\), the top-1 linkage
success rate is
\[
\mathrm{Link.}
\;\equiv\;
\mathrm{Link@1}
=
\frac{1}{N_{\mathrm{link}}}
\sum_{i=1}^{N_{\mathrm{link}}}
\mathbf{1}\!\left[\hat{c}_i=c_i^\star\right].
\]
This metric measures exact re-identification risk: whether the attacker's
highest-scoring candidate is the correct source person/record. It is directly
relevant here because the paper's \(k\)-anonymity gate is intended to make exact
record linkage ambiguous and therefore suppress successful top-1 matching.

\subsection{Privacy analysis and operating-point detail}
\label{app:privacy}

The privacy gate is fit on a training population of 2,662 quantized records. The selected operating point $(k,\ell)=(5,2)$ is applied to a prototype-support release surface containing 54 candidate support records, of which 32 remain visible and 22 are explicitly redacted.

Two patterns govern the frontier. First, under $\ell=1$, increasing $k$ yields the expected smooth privacy--utility trade-off: evidence utility decreases gradually as linkage exposure falls. Second, imposing $\ell=2$ is the decisive step for this release schema: once diversity is enforced, all tested values $k\in\{3,5,7,9\}$ collapse to the same operating regime. At $\ell=3$, the visible surface collapses entirely. The resulting picture is therefore not that privacy monotonically improves with every stronger setting, but that the diversity constraint is the binding variable on this release surface.

These release-surface diagnostics should be read together with the artifact-level attacks in the main paper. The supplement clarifies \emph{why} the frontier has the shape shown in Fig.~2, while the main paper's cross-method attack table evaluates the residual disclosure that remains once the selected release surface is turned into final exported evidence.

\section{Proofs of Theoretical Guarantees}
\label{supp:proofs}

This section provides the formal proofs for the theoretical guarantees established in Section 3 of the main text. We formalize the generative reporting process as a constrained Markov chain and leverage set-theoretic logic and probability theory to bound the test-time behavior of the Large Language Model (LLM).

\subsection{Proof of Theorem 1: Bounded Generative Faithfulness}

\textbf{Definition 1 (Atomic Claim Set).} Let $\phi: \mathcal{R} \to 2^{\mathcal{S}}$ be an extraction function that maps a natural language report $R \in \mathcal{R}$ to a set of discrete, atomic semantic claims $S_c \subset \mathcal{S}$. 

\textbf{Definition 2 (Strict Barrier Critic).} Let $\mathcal{C}_{E(x)} = \{R \in \mathcal{R} \mid \phi(R) \subseteq E(x)\}$ define the admissible constraint set of reports fully grounded in the evidence state $E(x)$. A critic energy function $E_{\text{critic}}(R)$ acts as a strict barrier if it assigns an infinite penalty to any report outside this set:
$$
E_{\text{critic}}(R) = 
\begin{cases} 
0, & \text{if } R \in \mathcal{C}_{E(x)} \\
\infty, & \text{otherwise}
\end{cases}
$$

\emph{Proof.} The test-time optimization loop operates over the grounded evidence state $E(x)$, which is completely defined by the deterministic visual and tabular differentials: $\Delta_j = (A(x)\cap A_j, A(x)\setminus A_j, A_j\setminus A(x), \Delta^{\text{tab}}_j)$.

Let $R^*$ be the accepted report returned by the Scribe-Critic search algorithm. By the definition of the algorithm's termination criteria, the report must satisfy $E_{\text{critic}}(R^*) \le \epsilon$. For a strictly deterministic barrier critic ($\epsilon = 0$), this implies $E_{\text{critic}}(R^*) = 0$.

Following Definition 2, $E_{\text{critic}}(R^*) = 0 \implies R^* \in \mathcal{C}_{E(x)} \implies \phi(R^*) \subseteq E(x)$. Because the comparative evidence state $E(x)$ is strictly bounded by $\Delta_j$, it holds that the atomic claims within the report are a strict subset of the pre-computed set-theoretic differentials. Therefore, the hypothesis space of the accepted report is mathematically restricted to the neuro-symbolic bottleneck, and the probability of the LLM introducing an ungrounded hallucinated comparison relative to this bottleneck is exactly $0$. $\blacksquare$

\subsection{Proof of Theorem 2: Information-Theoretic Privacy Bound}

To prove the privacy bounds, we model the framework as a Markov chain where information flows from the raw training prototype to the final generated report. We assume an adversary $\mathcal{A}$ whose prior belief matches the empirical distribution of the sanitized equivalence class (i.e., no auxiliary background knowledge linking identities to attributes).

\emph{Proof.} Let the flow of data be represented by the following sequence of random variables forming a Markov chain:
$$ S \to P \to \Sigma \to E \to R^* $$
where $S$ is a sensitive tabular attribute contained within the raw training prototype $P$, $\Sigma = \Psi(P)$ is the resulting non-differentiable sanitized semantic signature, $E$ is the grounded prompt context, and $R^*$ is the final generated report. Let lowercase letters denote their respective realizations (e.g., $S=s$, $R^*=r$).

Let $\mathcal{A}$ attempt to guess the sensitive attribute $S$ from the final report $R^*=r$. We seek to bound the maximum posterior probability $\max_{s} \Pr(S=s \mid R^*=r)$.

By the Law of Total Probability, we marginalize over all possible released signatures $\sigma$:
$$ \Pr(S=s \mid R^*=r) = \sum_{\sigma} \Pr(S=s \mid \sigma, R^*=r) \Pr(\sigma \mid R^*=r) $$

Because the system enforces a strict Markov chain, the raw attribute $S$ and the final report $R^*$ are conditionally independent given the bottleneck signature $\Sigma$. Therefore, $\Pr(S=s \mid \sigma, R^*=r) = \Pr(S=s \mid \sigma)$. Substituting this yields:
$$ \Pr(S=s \mid R^*=r) = \sum_{\sigma} \Pr(S=s \mid \sigma) \Pr(\sigma \mid R^*=r) $$

The privacy gate $\Psi$ enforces strict $\ell$-diversity on the visible signature $\sigma$, which guarantees that the empirical frequency of the most common sensitive attribute within any released equivalence class is at most $1/\ell$. Given $\mathcal{A}$ possesses no auxiliary knowledge to skew this distribution, $\max_{s} \Pr(S=s \mid \sigma) \le \frac{1}{\ell}$. Applying this bound:
$$ \Pr(S=s \mid R^*=r) \le \sum_{\sigma} \left( \frac{1}{\ell} \right) \Pr(\sigma \mid R^*=r) = \frac{1}{\ell} \sum_{\sigma} \Pr(\sigma \mid R^*=r) $$
Since $\sum_{\sigma} \Pr(\sigma \mid R^*=r) = 1$, we conclude:
$$ \max_{s} \Pr(S=s \mid R^*=r) \le \frac{1}{\ell} $$

Similarly, the $k$-anonymity constraint ensures that any released signature $\sigma$ corresponds to an equivalence class of at least $k$ distinct prototype records. By the identical Markov property argument, the posterior probability of an adversary successfully executing an exact record linkage attack (isolating a unique individual $i$ from $R^*$) is bounded by $\frac{1}{k}$. 

Because the generative model never observes $P$ directly, the artifact $R^*$ is mathematically secured against attribute inference and linkage attacks beyond these established thresholds. $\blacksquare$

\section{Extended clinical report gallery}
\label{app:gallery}

The main paper shows one final report per class. This section extends that view with additional same-class studies, restricted-comparison behavior, and harder threshold-sensitive examples. Each panel is written in the same final report voice as the main figure so that the supplement reads as an expanded gallery of outputs rather than as a case audit.

\small

\subsection{Additional routine reports}
The cases below extend the main figure with routine reports that stay clinically direct while varying in density reserve, structured risk burden, and comparison difficulty.

\begin{caseNormal}{Strongly preserved normal lumbar study}
\noindent \textbf{Query patient:} 86-year-old woman; lumbar DXA image plus tabular clinical record.
\par\smallskip
\textbf{ProtoMedX output:} \textsc{Normal}, $T=2.3$, Confidence 99.2\%; p1=\texttt{normal\_proto\_1} [N], p2/p3 = N/N.
\par\smallskip
\textbf{ProtoMedAgent report:} Lumbar bone density is well preserved and clearly within normal limits. The scan follows a strong preserved-density pattern, with the lower lumbar vertebrae standing out as denser than the upper spine and no convincing sign of generalized mineral loss. Mild degenerative change and small overlying artifact may accentuate that distal preservation, but they do not change the overall normal impression. Rheumatoid arthritis is the main structured risk factor in an otherwise light-burden profile. The study aligns with the high-preservation normal reference rather than any boundary osteopenic prototype.
\end{caseNormal}

\vspace{3pt}

\begin{caseOsteopenia}{Early osteopenia with low non-age risk burden}
\noindent \textbf{Query patient:} 95-year-old woman; lumbar DXA image plus tabular clinical record.
\par\smallskip
\textbf{ProtoMedX output:} \textsc{Osteopenia}, $T=-1.0$, Confidence 94.2\%; p1=\texttt{osteopenia\_proto\_0} [Op], p2/p3 = Op/Op.
\par\smallskip
\textbf{ProtoMedAgent report:} Low bone mass consistent with early osteopenia is present. The estimate sits at the osteopenic threshold, and the scan no longer shows the smooth preserved-density appearance expected of a normal lumbar spine. Some lower-lumbar reserve remains, but the overall impression is one of mild distributed attenuation rather than advanced depletion. Advanced age keeps the finding clinically relevant even without additional major structured risk factors. The nearest prototype comparison is an early-loss osteopenic reference rather than a truly preserved normal card.
\end{caseOsteopenia}

\vspace{3pt}

\begin{caseOsteoporosis}{Advanced osteoporosis with prior fragility fracture}
\noindent \textbf{Query patient:} 84-year-old woman; lumbar DXA image plus tabular clinical record.
\par\smallskip
\textbf{ProtoMedX output:} \textsc{Osteoporosis}, $T=-4.0$, Confidence 97.7\%; p1=\texttt{osteoporosis\_proto\_2} [Os], p2/p3 = Op/Op.
\par\smallskip
\textbf{ProtoMedAgent report:} Bone density is in the osteoporotic range with advanced mineral loss. The estimate lies far below the osteopenic boundary, and the lumbar spine reads as diffusely depleted rather than residually borderline. Previous fragility fracture places the patient in a high-risk context for future fracture. The study is best matched to a severe low-density osteoporotic reference and is clearly separated from any early-loss osteopenic prototype.
\end{caseOsteoporosis}

\subsection{Edge-Case reports}
These reports show that the final report style remains usable when only part of the comparison surface is visible or when the density evidence sits close to a diagnostic threshold.

\begin{caseNormal}{Near-threshold normal study under restricted comparison release}
\noindent \textbf{Query patient:} 88-year-old woman; lumbar DXA image plus tabular clinical record.
\par\smallskip
\textbf{ProtoMedX output:} \textsc{Normal}, $T=-0.2$, Confidence 86.5\%; p1=\texttt{normal\_proto\_1} [N], p2/p3 = N/Op.
\par\smallskip
\textbf{ProtoMedAgent report:} Lumbar bone density remains within normal limits. The estimate lies close to the normal/osteopenia boundary, but the visible comparison surface still favors preserved density over early mineral loss. The study retains a preserved lower-lumbar-greater-than-upper pattern rather than the flatter, weaker appearance of an osteopenic reference. Previous fragility fracture keeps the case clinically important despite the normal density impression, so the report remains normal in class while using careful near-threshold wording.
\end{caseNormal}

\vspace{3pt}

\begin{caseOsteopenia}{Borderline osteopenia with fracture and rheumatoid arthritis}
\noindent \textbf{Query patient:} 74-year-old woman; lumbar DXA image plus tabular clinical record.
\par\smallskip
\textbf{ProtoMedX output:} \textsc{Osteopenia}, $T=-0.6$, Confidence 61.2\%; p1=\texttt{osteopenia\_proto\_0} [Op], p2/p3 = Op/N.
\par\smallskip
\textbf{ProtoMedAgent report:} Early low bone mass is favored despite a near-normal estimated T-score. The calibrated interval crosses the normal/osteopenia boundary, and the scan reads closer to an early mineral-loss pattern than to a convincingly preserved normal spine. Previous fragility fracture and rheumatoid arthritis make this borderline presentation clinically significant. Comparison with lower-boundary osteopenic references therefore supports cautious osteopenic wording and follow-up rather than dismissal as preserved density.
\end{caseOsteopenia}

\vspace{3pt}

\begin{caseOsteoporosis}{Low-density study straddling osteopenia and osteoporosis}
\noindent \textbf{Query patient:} 85-year-old woman; lumbar DXA image plus tabular clinical record.
\par\smallskip
\textbf{ProtoMedX output:} \textsc{Osteoporosis}, $T=-2.0$, Confidence 61.8\%; p1=\texttt{osteoporosis\_proto\_2} [Os], p2/p3 = Op/Op.
\par\smallskip
\textbf{ProtoMedAgent report:} Low bone density is present, but the final impression remains clinically measured. Image features raise concern for osteoporosis, yet the calibrated estimate remains in the osteopenic range and spans the osteopenia/osteoporosis threshold. The visible comparison neighborhood is weaker than preserved or early-loss normal references, but it still does not reproduce the most depleted severe osteoporotic patterns across the entire lumbar column. With age as the main risk driver and no stronger structured burden, correlation with prior DXA studies or interval follow-up is appropriate before labeling the scan unequivocal advanced osteoporosis.
\end{caseOsteoporosis}

\small

\section{Prototype atlas (ProtoCards)}
\label{app:prototypes}

The main paper highlights three reference ProtoCards. The atlas below organizes the full learned library by diagnostic class and makes explicit what each prototype contributes: the density regime it occupies, the scan pattern that tends to recur, the surrounding clinical context, and the caveat that most often changes interpretation.

\small

The atlas below enumerates the full learned prototype library in the same narrated style as the main figure. Each card is written as a usable clinical reference: what the lumbar scan tends to look like, which clinical backdrop tends to recur, and why that prototype matters when it is contrasted against neighboring cards.

\subsection{Normal reference prototypes}
The normal atlas ranges from strongly preserved spines to cautious near-threshold normal studies in which fracture history or inflammatory context can coexist with preserved density.

\begin{caseNormal}{Heterogeneous mild preserved-density normal prototype}
\textit{Reference prototype: \texttt{normal\_proto\_0}.}

Typical study: density remains preserved overall, but the lumbar column is visibly uneven from vertebra to vertebra. Lower lumbar vertebrae often read slightly denser than the upper spine, while unreadable labels, incomplete coverage, and mild lower-lumbar degenerative change repeatedly force measured wording instead of an emphatically pristine normal report.
\end{caseNormal}

\vspace{4pt}

\begin{caseNormal}{Robust preserved-density normal prototype}
\textit{Reference prototype: \texttt{normal\_proto\_1}.}

Typical study: the lower lumbar spine remains distinctly denser than the upper lumbar vertebrae and the whole column reads comfortably normal across a broad preserved range. Prior fragility fracture can coexist with this prototype, but unlike the boundary normal or osteopenic references the preserved structure remains convincing even when mild sclerosis or overlay exaggerates the distal vertebrae.
\end{caseNormal}

\vspace{4pt}

\begin{caseNormal}{Mid-range preserved-density normal prototype}
\textit{Reference prototype: \texttt{normal\_proto\_2}.}

Typical study: density is clearly preserved but without the excess reserve of the strongest normal anchor. This prototype fits quieter normal scans, usually with lighter lifestyle or body-composition risk context and fewer recurring artifact cues than the more heterogeneous normal references.
\end{caseNormal}

\vspace{4pt}

\begin{caseNormal}{Boundary-side normal prototype with fracture context}
\textit{Reference prototype: \texttt{normal\_proto\_3}.}

Typical study: the scan stays on the preserved side of the normal/osteopenia boundary, but the surrounding clinical context is heavier than in the tighter normal references. Prior fragility fracture recurs often enough that this prototype is useful when clinical risk is elevated even though the lumbar density pattern still reads normal.
\end{caseNormal}

\vspace{4pt}

\begin{caseNormal}{Tight near-threshold normal prototype with inflammatory risk}
\textit{Reference prototype: \texttt{normal\_proto\_4}.}

Typical study: density remains technically normal, but only with a small reserve above the osteopenic boundary. Low BMI, fracture history, or inflammatory context often make this a cautious preserved-density report rather than a strongly reassuring one.
\end{caseNormal}

\vspace{4pt}

\begin{caseNormal}{Compact mild-normal prototype with fracture history}
\textit{Reference prototype: \texttt{normal\_proto\_5}.}

Typical study: a mild preserved-density pattern centered close to the boundary of normality, without the stronger reserve of the high-preservation references. Everyday fracture or lifestyle context may still accompany this prototype, so it supports restrained normal wording rather than an emphatic statement of high skeletal reserve.
\end{caseNormal}

\subsection{Osteopenic reference prototypes}
The osteopenic atlas is organized around lower- versus upper-boundary low bone mass, with additional separation between fracture-linked, lifestyle-linked, and especially cohesive early-loss patterns.

\begin{caseOsteopenia}{Early mineral-loss osteopenic prototype with vertebral heterogeneity}
\textit{Reference prototype: \texttt{osteopenia\_proto\_0}.}

Typical study: lower lumbar vertebrae may retain relatively greater density, but preservation becomes uneven across the visible lumbar column and the upper lumbar spine shows reduced density relative to the lower vertebrae. Rotation, limited coverage, unreadable vertebral labels, degenerative change, and focal overlay recur often enough that the report should sound measured, yet the prototype still reads clearly weaker than a normal spine.
\end{caseOsteopenia}

\vspace{4pt}

\begin{caseOsteopenia}{Tight low-burden lower-boundary osteopenia}
\textit{Reference prototype: \texttt{osteopenia\_proto\_1}.}

Typical study: real but subtle low bone mass clustered near the lower osteopenic boundary, with little dramatic visual flourish. This prototype is useful when the class decision comes from a consistent borderline deficit rather than a memorable structural cue, often in patients with leaner body habitus or prior fracture.
\end{caseOsteopenia}

\vspace{4pt}

\begin{caseOsteopenia}{Broader fracture-linked lower-boundary osteopenia}
\textit{Reference prototype: \texttt{osteopenia\_proto\_2}.}

Typical study: a lower-boundary osteopenic scan that stays close to normal on density alone, but appears in a clinically heavier fracture-linked context. The visual family is mixed rather than tidy, so this prototype supports osteopenic wording for heterogeneous early-loss presentations rather than textbook boundary examples.
\end{caseOsteopenia}

\vspace{4pt}

\begin{caseOsteopenia}{Lifestyle-linked lower-boundary osteopenia}
\textit{Reference prototype: \texttt{osteopenia\_proto\_3}.}

Typical study: mild lower-boundary osteopenia accompanied more often by lifestyle-associated risk, especially heavier alcohol exposure, than by recurrent fracture. The density loss is gentle and the image family broad, making this a useful prototype for clinically lighter osteopenia that still deserves recognition as low bone mass.
\end{caseOsteopenia}

\vspace{4pt}

\begin{caseOsteopenia}{Upper-boundary osteopenia near preserved density}
\textit{Reference prototype: \texttt{osteopenia\_proto\_4}.}

Typical study: a subtly weakened lumbar spine that leans toward preserved density more than any other osteopenic reference. This prototype is most helpful when a study nearly passes as normal but still shows enough diffuse weakening to justify osteopenic wording.
\end{caseOsteopenia}

\vspace{4pt}

\begin{caseOsteopenia}{Cohesive fracture-associated lower-boundary osteopenia}
\textit{Reference prototype: \texttt{osteopenia\_proto\_5}.}

Typical study: definite early mineral loss with one of the most stable image patterns in the osteopenic class. Previous fragility fracture may recur, but the main value of this prototype is that it gives a crisp visual anchor when a study is clearly osteopenic even though it is not yet osteoporotic.
\end{caseOsteopenia}

\subsection{Osteoporotic reference prototypes}
The osteoporotic atlas separates moderate versus deeper low-density anchors and distinguishes relatively isolated density loss from the highest-burden severe disease.

\begin{caseOsteoporosis}{Moderate low-density osteoporosis with residual lower-lumbar contrast}
\textit{Reference prototype: \texttt{osteoporosis\_proto\_0}.}

Typical study: overall osteoporotic density loss with residual lower-lumbar-greater-than-upper contrast still visible. That apparent preservation must be read cautiously because degenerative change, overlay, and incomplete coverage recur here; this prototype fits scans that are osteoporotic overall but not uniformly depleted across all visible vertebrae.
\end{caseOsteoporosis}

\vspace{4pt}

\begin{caseOsteoporosis}{Broad severe osteoporosis with secondary-cause context}
\textit{Reference prototype: \texttt{osteoporosis\_proto\_1}.}

Typical study: unmistakably low density, but with mixed secondary-cause context and a visually diverse severe support set. This prototype is most useful when advanced skeletal fragility is clear even though no single picture-perfect osteoporotic template dominates the presentation.
\end{caseOsteoporosis}

\vspace{4pt}

\begin{caseOsteoporosis}{Deep low-density osteoporotic prototype}
\textit{Reference prototype: \texttt{osteoporosis\_proto\_2}.}

Typical study: deeper low density that sits farther from the osteopenic boundary than the moderate severe references. The defining feature is depth of mineral loss rather than a dense accompanying risk profile, so the image itself carries most of the diagnostic weight.
\end{caseOsteoporosis}

\vspace{4pt}

\begin{caseOsteoporosis}{High-burden severe osteoporosis prototype}
\textit{Reference prototype: \texttt{osteoporosis\_proto\_3}.}

Typical study: marked low density accompanied by the richest clinical burden in the atlas, including recurrent fracture, smoking, and inflammatory risk. This prototype represents the part of the osteoporotic class in which both the scan and the clinical history point strongly toward advanced skeletal fragility.
\end{caseOsteoporosis}

\vspace{4pt}

\begin{caseOsteoporosis}{Tight deep-low-density osteoporotic subset}
\textit{Reference prototype: \texttt{osteoporosis\_proto\_4}.}

Typical study: a narrow deep-low-density subset in which the T-score band is extremely severe even when other structured risks are relatively sparse. This prototype is the right reference when the scan is unequivocally osteoporotic on density alone and the report should emphasize severity rather than risk accumulation.
\end{caseOsteoporosis}

\vspace{4pt}

\begin{caseOsteoporosis}{Cohesive low-density osteoporosis with prior-fracture motif}
\textit{Reference prototype: \texttt{osteoporosis\_proto\_5}.}

Typical study: consistently depleted lumbar mineralization with recurrent prior fragility fracture across a very cohesive severe support set. This prototype provides the cleanest decisive comparison when a case is clearly beyond early low bone mass and no longer sits in a threshold-sensitive regime.
\end{caseOsteoporosis}


\end{document}